# Exploiting multilingual nomenclatures and language-independent text features as an interlingua for cross-lingual text analysis applications

**Ralf Steinberger, Bruno Pouliquen & Camelia Ignat**

European Commission – Joint Research Centre (JRC)
Via E. Fermi, T.P. 267, 21020 Ispra (VA), Italy
Firstname.Lastname@jrc.it – http:// www.jrc.it/langtech

**Abstract**
We are proposing a simple, but efficient basic approach for a number of multilingual and cross-lingual language technology applications that are not limited to the usual two or three languages, but that can be applied with relatively little effort to larger sets of languages. The approach consists of using existing multilingual linguistic resources such as thesauri, nomenclatures and gazetteers, as well as exploiting the existence of additional more or less language-independent text items such as dates, currency expressions, numbers, names and cognates. Mapping texts onto the multilingual resources and identifying word token links between texts in different languages are basic ingredients for applications such as cross-lingual document similarity calculation, multilingual clustering and categorisation, cross-lingual document retrieval, and tools to provide cross-lingual information access.

## 1. Background and Motivation

The European Union (EU) currently has 20 official languages, plus a few non-official ones. Most existing text analysis software tools have been developed for a few major languages, while very few resources and tools are available for the less widely spoken languages. There clearly is a need for more tools that can help the European citizens to access textual information written in the other languages.

The 20 official EU languages add up to 190 language pair combinations. Almost all *cross-lingual* text analysis applications, including Machine Translation (MT), Cross-Lingual Information Retrieval (CLIR) and Cross-Lingual News Topic Tracking (CLNTT), make use of *bilingual* equivalences and rules. The few approaches to CLNTT, for instance, are either based on bilingual dictionaries (Wactlar 1999) or use MT (Leek et al. 1999). In the EU setting, interlingua approaches and approaches towards unified multilingual resources, such as EuroWordNet and MULTEXT, clearly gain in attraction. However, there are many more unexploited resources that may not have been developed for machine use, but that can be exploited for multilingual Information Extraction (IE) and to provide cross-lingual information access.

The Language Technology team of the *Joint Research Centre* (JRC) has the aim to produce a number of text analysis applications for ideally all official EU languages (and more) that help users to navigate in large multilingual document collections and that provide them with cross-lingual information access. Due to a lack of manpower and due to the limited availability of machine-usable linguistic resources, we developed the following preferences:
(a) limiting language-specific text processing to a minimum, by using heuristics and other shallow methods;
(b) using statistics and Machine Learning (ML) methods rather than hand-crafted linguistic rules, where possible;
(c) making use of various available multilingual lexical resources, even if they were not initially developed for machine use.
While it is clear that more thorough knowledge-driven methods would produce better results in many cases, the JRC's work has shown that a shallow and mostly language-independent approach can yield a number of useful and new text analysis applications while keeping the language-specific effort to between one and three person months of effort per language.

The following sections describe efforts to map texts onto multilingual knowledge structures (Section 2) and to exploit further almost language-independent text features (Section 3). Section 4 explains how to deal with some language-specific issues and Section 5 lists a few language-independent methods and tools that can be used together with the resources mentioned in the previous sections. Section 6 shows some useful applications built with the procedures described in this article. In Section 7, we draw a few conclusions.

## 2. Mapping texts onto existing multilingual thesauri, nomenclatures and gazetteers

When mapping a given text onto a knowledge structure such as a thesaurus, we create a text representation consisting of a choice of thesaurus nodes, and possibly also of the relative importance of various nodes for the text representation. One, but not the only way of carrying out this mapping process is by verifying the lexical overlap between the document's vocabulary and the terms of the thesaurus. Two documents can be assumed to be similar if they have a similar representation according to the mapping onto this thesaurus.

In a multilingual thesaurus, nodes in the various language versions are linked via language-independent (typically numerical) node identifiers. While the conceptual world of a given language or of a specific thesaurus is, of course, not completely language-independent, the numerical thesaurus links between various language versions are good enough for an interlingua approximation. Two documents written in different languages can thus be assumed to be similar if they have a similar text representation according to this multilingual thesaurus.

Additionally to thesauri, gazetteers and nomenclatures can fulfil the same function. *Gazetteers* are geographical dictionaries, i.e. lists of place names. According to Norviliené (forthcoming), the term *nomenclature* is used to describe ordered systems of words (e.g. product names) used in a particular discipline (e.g. business or customs), containing a description of entities from a particular domain and their, typically mono-hierarchical, relationship. *Thesauri* are poly-hierarchically ordered systems of concepts and their natural language names that are mainly used for documentation purposes such as indexing and retrieval.

**Figure 1.** Recognition of place names in Bulgarian and Czech text, and display of the results in English.

The aim of this section is to show how texts can be mapped onto one or more thesauri to create a multi-faceted language-independent document representation. The more thesauri can be used, the more information will be available for the document representation and the better documents can be compared with each other. The following sub-sections sketch our current mapping process onto various such lexical knowledge sources.

### 2.1. Gazetteers of place names

Unlike people's names and other named entities, place names cannot be recognised by searching for patterns in text because there are as good as no contextual clues (Gey 2000). Instead, geographical place name recognition has to rely on gazetteers and can only be carried out via a lookup of text words in the gazetteer. As places are spelled with a first uppercase letter in EU languages, only uppercase words need to be looked up. The lookup process sounds simple, but there are four major difficulties:
(a) Place names can also be words in one or more languages, such as 'And' (Iran) and 'Split' (Croatia);
(b) Some place names are homonymic with people's names, such as 'Victoria' (capital of the Seychelles, and others) and 'Annan' (UK);
(c) Many major places have varying names in different languages (exonyms; Venezia vs. Venice, etc.) or even in the same language ('Saint Petersburg', 'Saint Pétersbourg', '**Санкт-Петербург**' [Sankt-Peterbúrg], 'Leningrad', 'Petrograd', etc.).
(d) Multiple places share the same name, such as the fourteen cities and villages in the world called 'Paris';

While place name recognition in general is a very well understood named entity recognition task, disambiguation between various homographic place names (issue d) has only recently been tackled (Pouliquen et al. 2004a). Exonym recognition (issue c) has to rely on an exhaustive multilingual database. While a number of monolingual gazetteers are freely available (see Gey 2000) we are only aware of two multilingual place name lists: the KNAB database of the Institute of the Estonian Language[1] and the European Commission's NUTS database[2], currently available in fifteen languages.

Even for languages with relatively few speakers such as Slovene, good resources exist. For instance, KNAB currently contains about 150 Slovene place names. The freely available database of the *Geonet Name Server*[3] has 6600 English language references of Slovene place names. Slovene place names are handled by the *Slovene governmental commission for the standardisation of geographical names*[4], who even provide a link to a gazetteer of exonyms[5].

Our approach consists of looking up all uppercase words in the gazetteer database and of applying a number of heuristics for disambiguation (see Pouliquen et al. 2004a). When a string could be a single word or be part of a multi-word place name, the longer place name is preferred. The result is a list of place names occurring in the text with their offset and length, plus latitude and longitude, as well as information on the country they belong to and probably information about the hierarchical organisation of the country (e.g. town, province, region, country). In Figure 1, automatically identified place names in Bulgarian and Czech text are highlighted and translated. Additional information is available in the underlying XML file, but is not displayed here.

To limit the negative impact of place names that could also be common words or people's names (problems (a) and (b)), which would lead to many wrong hits and thus to a low precision, we currently use lists of *geo-stop words*, i.e. words that should not be marked as place names even if they are found in text. As ambiguous place names such as 'And' and 'Split' are only a problem for English language texts, but not for German or other languages, there should be a different geo-stop word list for each language. Producing a geo-stop word list for a new language takes little effort as word frequency lists of the language can be used. By automatically geo-coding a frequency list of the ten thousand most frequent words of the language and collecting those words that were found by the system, but that are not place names, such a geo-stop word list is quickly produced. Person names such as *Victoria* are harder to come around as the person name is rather frequent and there are 190 places with this name in the world, including the capital of the Seychelles. This problem can only be overcome by using the outcome of the person name recognition tool described in section 3.2.

An evaluation of the place name recognition tool in English texts yielded a precision of 96.8% and a recall of 96.5%. For details, see Ignat et al. (2003).

Language-specific issues regarding the lookup process, such as place name inflection, will be discussed in section 4 as they are not only relevant to place name recognition.

The result of the mapping process is thus a vector of place names where each place name is a dimension and the frequency with which it has been mentioned in the text is the length of the vector. For some applications, it may be useful to restrict the recognition resolution to the country level, i.e. each mention of a place in the country adds to the country score. The occurrence frequency and the country score can also be normalised, using TF.IDF or similar, to down-weight the importance of places like *Washington* that are highly frequent in some text types such as world news.

---

[1] See http://www.eki.ee/knab/knab.htm.
[2] Available at http://europa.eu.int/comm/eurostat/ramon/.
[3] See http://earth-info.nga.mil/gns/html/.
[4] See http://www.sigov.si/kszi/.
[5] Available at http://www.sigov.si/kszi/ang/exonyms.pdf.

| TARIC CODE | PRODUCT DESCRIPTION |
|---|---|
| 0702 | Tomatoes, fresh or chilled |
| 0702 00 00 07 | Cherry tomatoes |
| 0702 00 00 99 | Other |
| 0703 | Onions, shallots, garlic, leeks and other alliaceous vegetables, fresh or chilled |
| 0703 10 | Onions and shallots |
| 0703 20 | Garlic |
| 0703 90 | Leeks and other alliaceous vegetables |
| 0703 90 00 10 | Leeks |
| 0703 90 00 90 | Other |

**Table 1.** English product descriptions in TARIC chapter: *Edible Vegetables and Certain Roots and Tubers*.

**2.2.　Nomenclatures of products, etc.**

Other views of the same document can be produced by listing all document terms from various other fields, such as products and product groups, professions, medical or electro-technical terms, etc. Various nomenclatures can be downloaded from the internet (see Norvilienè 2004), and many of them are available on the EC's classification server *Ramon* (see Footnote 2). For instance, there is the electro-technical nomenclature ETIM[6], the *Statistical Classification of Products by Activity in the European Economic Community* CPA, the *Statistical Classification of Economic Activities in the European Community* NACE, and many more.

To date, we have only worked with the *Integrated Tariff of the European Communities* TARIC[7], which is the hierarchical product list used by the Customs Offices in the EU to declare the movement of goods across borders. TARIC is a more detailed version of the so-called *Combined Nomenclature* CN, which is again more detailed than the *Harmonised System* HS used by the World Customs Organisation. TARIC distinguishes about 28,000 headings and subdivisions.

We chose TARIC because it exists in twenty languages (including Slovene) and because it is a rather complete list of tangible items that can be imported or exported. It is in the nature of TARIC that illegal products such as *bombs* and many drugs are not included (although *heroin* and *cocaine* are part of TARIC). It includes live animals, food, chemicals, pharmaceuticals, textiles, precious stones, metals, machinery, vehicles, optical material, works of art, and much more.

Table 1 shows some of the product descriptions that are organised hierarchically into up to 5 levels (two digits per level). Knowing which of the products and product groups are referred to in a text can be very useful to generate a product-related document representation, i.e. a vector of products and their relative importance in the text. We can furthermore use the numerical TARIC codes as an interlingua to represent the product aspect of document written in the twenty languages in which the product nomenclature exists. However, before being able to use the product lists of this resource in a lookup process, we needed to overcome several difficulties:

(a) As the entire TARIC product description (e.g. "*Leeks and other alliaceous vegetables*" in code 070390) will not be found verbatim in the text, the product terminology first needs to be extracted from the description (e.g. *leeks* and *alliaceous vegetables* in Table 1).

(b) Usually, the plural forms are used in TARIC so that the singular or other inflected forms need to be added for the lookup process to be successful. For further issues concerning inflection of words and suffixes, see Section 4.

(c) Syntactic co-ordination constructions such as in code 0703 need to be resolved and expanded out to produce lists such as *fresh onions*, *chilled onions*, *fresh shallots*, *chilled alliaceous vegetables*, etc.

(d) This process typically results in product lists such as *fresh onions* and *chilled onions*, while the most usual underspecified term *onions* is not part of the list. This needs to be added.

(e) While multi-word terms are usually monosemous, many single-word terms such as *onion* or *juice* can be part of many different TARIC classes as there are many different types of juices and onions (*wild onions*, *pearl onions*, *dried onions*, etc.). As we did not want to miss frequently used products such as *onions* or *juice*, and we did not want one term to trigger many different TARIC classes, we decided to add about 350 super-groups such as *vegetables* and *milk products* and to place the under-specified term directly under the super-group.

These steps were carried out, mostly by the *Centre for Information and Language Processing* CIS[8] at the University of Munich in Germany, in the context of a collaborative agreement, for the languages English, German, French, Spanish, Italian and Portuguese. In the semi-automatic process, heuristics were used and results were checked manually. Inflection forms were added by making use of extensive morphological dictionaries available at CIS. The English and Italian dictionary resources created by CIS were then checked thoroughly for correctness at the JRC.

The resulting dictionaries are thus of the form SUPER-GROUP | CODE | TERM where several terms are allowed for the same code if written one term per line, and several codes are obviously allowed for each super-group. The super-group column furthermore allows us to do a more coarse-grained classification of texts so that documents triggering the class *vegetables* several times are identified as similar even if they do not mention the same vegetables. To date, the dictionaries have been developed for the languages English, Italian, German, French, Spanish and Portuguese.

Regarding the recognition of the derived product terminology in the text, the same lookup procedure can be used as for geographical place names. However, in most European languages, products are not spelled with a first uppercase letter so that all words need to be checked against the terms in the product list. Figure 2 shows some product recognition results.

---

[6] See http://www.etim.de/html/download.html
[7] See http://europa.eu.int/comm/taxation_customs/databases/taric_en.htm
[8] See http://www.cis.uni-muenchen.de/

The difficulties involved in the lookup process are again linked to polysemous words like *bush*, *joint*, *bus*, etc. Some of these terms belong to very different TARIC classes (e.g. *joint*). Others are simply homographic with words not related to products (e.g. *Bush*). For testing, we applied the system to various text types and, more importantly, to the 10,000 top frequent words derived from reference corpora. This gave us a good idea of the most frequent missing products, which were then added to the dictionaries. Furthermore, this helped us to identify those high-frequency words that are homographic with products and that could thus potentially generate wrong hits. Depending on the type of problem, we used one of two solutions. (a) For words triggering different TARIC product classes, we usually amended the dictionary by adding some additional specification (e.g. *joint* was changed to *rubber joint*) that helps in the disambiguation. The disadvantage is that the single word *joint* will no longer be recognised. (b) For words that are homographic with non-product vocabulary of the language (e.g. *Bush*), we produced a language-dependent *product stop word list* containing all those words that the system should not recognise. This helps to avoid that the US president triggers the product class *live plants*. We thus decided to sacrifice recall for precision.

The effort to prepare and tune the product dictionaries for each language ranges between two and six months per language, but we foresaw that the advantage of mapping texts onto the TARIC nomenclature with its encompassing coverage would be worth the effort. The result of the product recognition procedure is thus a product information extraction tool that allows us also to provide users with product-specific cross-lingual information access and to produce a product-specific feature vector for each document that can be used for monolingual and cross-lingual document similarity calculation.

The TARIC nomenclature is seemingly distributed for free, but the dictionaries derived from it cannot currently be made available due to the agreement with CIS. However, the JRC would be interested in collaborations creating publicly available resources for more languages.

### 2.3. Thesauri and classification systems

Libraries and documentation centres of most large organisations use hierarchically organised thesauri or flat lists of subject domain descriptions as classification systems to store and retrieve their documents. Documents are often multiply classified, meaning that each document is marked as belonging to several classes (*multi-label categorisation*). Such a classification of a document leads to yet another vector space representation of documents, using the descriptors as dimensions and, if the descriptors are ordered or weighted, the weight as vector length.

| They ate young river salmon with cream and potatoes. |||
|---|---|---|
| (Milk product)-EN 0401000000 80 | cream, milk | leite, nata |
| (vegetable)-EN 0710100000 80 | potatoe, potatoes | batatas |
| (fish)-EN 0301991910 80 | young river salmon | alivius de salmío |

**Figure 2.** Automatic recognition of products in English text. Display of the results in English and Portuguese.

The European Parliament (EP) and the European Commission (EC) have jointly developed a thesaurus called EUROVOC (EUROVOC 1995) that is used by them and about twenty regional and national European parliaments to index (i.e. classify) their texts. Though other classification systems exist, EUROVOC is adapted by a growing number of national organisations so that it has now become sort of a standard. To obtain a licence, it is necessary to contact the EC's Publications Office OPOCE.

EUROVOC is a wide-coverage thesaurus that organises its over 6,000 descriptors (classes) from 21 different *fields* (e.g. *politics*, *finance*, *science*, *social questions*, *organisations*, *foodstuff*, etc.) hierarchically into a maximum of 8 levels. EUROVOC exists in currently 22 languages where each numerical descriptor code has exactly one terminological correspondence per language.

As EUROVOC is a wide-coverage thesaurus with only 6000 classes, its descriptors are mostly rather high-level, conceptual terms. Examples are PROTECTION OF MINORITIES, FISHERY MANAGEMENT and CONSTRUCTION AND TOWN PLANNING.[9] Unlike the concrete low-level terms from TARIC and many other nomenclatures, EUROVOC descriptors cannot normally be *extracted* from texts, i.e. they can only rarely be found via a lookup procedure. Instead, EUROVOC classification is a keyword *assignment* task, i.e. the most pertinent descriptors from an independent reference list (the thesaurus) are assigned to a text even if these terms do not occur verbatim in the text.

In the various European parliaments, this assignment is done manually by professional librarians, but the JRC has developed a system that learns from manually classified documents to assign a ranked list of EUROVOC descriptors to any given text. This work is described in detail in Pouliquen et al. (2003a) so that we only summarise the procedure here: The system maps documents onto EUROVOC by carrying out category-ranking classification using Machine Learning methods. In an inductive process, it builds a profile-based classifier by observing the manual classification on a training set of documents with only positive examples. Table 2 shows the first few of a long list of words automatically identified as being significant for the EUROVOC descriptor FISHERY MANAGEMENT. Before feeding the training texts to the ML algorithm, some linguistic pre-processing was carried out to lemmatise words and to mark up multi-word terms such as *power_plant* and *New_York* as one token and a large stop word list of words with low semantic content was used. However, tests have shown that lemmatisation and multi-word mark-up had only little impact on the performance for Spanish and English. Assignment results for the highly inflected Finnish language were very comparable, showing that the statistical method can be applied without using linguistic tools, if necessary.

A manual evaluation of the EUROVOC descriptor assignment process for English and Spanish parliamentary documents, taking human performance as a benchmark, showed that the system performs 86% and 80% as well as the professional indexers did. For details, see Pouliquen et al. (2003a).

The outcome of the mapping process for a given text is a ranked list of the EUROVOC classes that are most pertinent for this text. Table 3 shows the first few EUROVOC

---
[9]We write all EUROVOC descriptors in small caps.

| Lemma | Weight |
|---|---|
| fishery_resource | 54.47 |
| fishing | 49.11 |
| fish | 46.19 |
| common_fishery_policy | 44.67 |
| fishery | 44.19 |
| fishing_activity | 43.37 |
| fly_the_flag | 42.87 |
| aquaculture | 39.27 |
| conservation | 38.34 |
| vessel | 37.91 |

**Table 2**. The first few of a long list of lemmas that have been automatically identified as being highly relevant and typical for documents that were manually classified with the EUROVOC descriptor FISHERY MANAGEMENT, plus their weight (the *profile* of the descriptor). The presence of many of these lemmas in a given text indicate a certain likelihood that FISHERY MANAGEMENT is an appropriate descriptor for this text.

descriptors assigned automatically to a text found on the internet.

Due to the multilingual nature of EUROVOC, this representation is independent of the text language so that it is very suitable for cross-lingual document similarity calculation. The system has currently been trained for thirteen languages so that documents written in any of these languages can be represented with the same language-independent EUROVOC descriptor vector. Unlike the applications described in sections 2.1 and 2.2, this Machine Learning method to map documents onto thesauri requires training material, i.e. documents that have been manually classified. While some linguistic, rule-based or dictionary-based approaches exist for automatic thesaurus indexing (e.g. Marjorie & Hainebach 1996), more recent efforts such as the one by Montejo-Ráez (2002) tend to exploit the power of ML approaches. The advantage of these becomes even more evident for highly multilingual applications such as automatic EUROVOC indexing.

Most other highly multilingual thesauri we are aware of are subject-specific, such as the agricultural thesaurus AGROVOC, the particle physics thesaurus DESY and the medical thesauri UMLS and MeSH. AGROVOC, which is freely available at the FAO web site[10], exists in six major world languages. The medical thesaurus MeSH exists in twelve mainly European languages, but according to Nelson et al. (2000), the thesaurus has fully or partially been translated into a further eight world languages, including Slovene.

### 3. Language-independent text features

The mapping processes described in section 2 yield several vector space document representations, one for each thesaurus, nomenclature, gazetteer or word list used. Further multilingual representations can be generated by extracting named entities to create lists of text features such as (a) date or (b) currency expressions, (c) numbers and (d) names, as these can be represented in a normalised, language-independent format. For an introduction to the state of the art of the field of Named Entity Recognition (NER), see Daille & Morin (2000). Names of people or organisations are not strictly language-independent because names may be written differently depending on the language (and sometimes even within the language), but at least among European languages many names are spelled the same. Due to the historical relatedness of many European languages, there are even (e) a few general language words that are similar or the same. These are usually referred to as *cognates*. The English and German words 'finger', 'arm', 'demonstration', 'computer', etc. are some examples. In this section, we describe how these five additional text features can be recognised and exploited to contribute to linking related documents both monolingually and across languages.

### 3.1. Date and currency expressions

Within the same language, there are usually different ways of writing a certain date or currency expression (e.g. English *13 October 2004, 13/10/2004, 13.10.2004, thirteenth of October of the year two thousand and four, etc.*). Some of these date expressions may be the same as in other languages (e.g. 13.10.2004), but others are not. As the underlying concept is the same, namely a reference to a specific date in the same time reference system, the concept can be expressed in a standard way (see, for instance, ISO standard ISO-8601) so that it is the same across languages. For dates, we currently use 'DD'YYYYMMDD. Expressions such as 13.10.2004 are thus normalised to DD20041013.

At the JRC, we do not currently recognise currency expressions, but we have developed a tool that recognises and normalises date expressions. It is a language-independent software tool that uses language-specific parameter files, one per language. The set of languages includes Slovene. A preliminary version of this tool is described in Ignat et al. (2003). It is available on request.

The language-specific parameter file allows to list days of the week, months of the year, common abbreviations for week days and months, cardinal and ordinal number expressions, words that can be part of the date expression (e.g. *of the year*), as well as expressions used for relative dates such as *yesterday*, *last December*, etc. It furthermore allows to specify ordering rules. In English, for instance, it is possible to mention the DAY after the MONTH (e.g. *May 2nd*) whereas this is not allowed in German and other languages. The tool recognises *absolute* and *relative* dates, as well as *complete* and *incomplete* dates. The expression *last December* thus is a relative incomplete date with underspecified DAY. If a *reference date* is given (this can, for instance, be the publication date for newspaper articles), the tool can calculate the normalised expression DD20031200 for the words *last December* if the reference date is in the year 2004.

---

[10] See http://www.fao.org/agrovoc/.

| Rank | Descriptor | Similarity |
|---|---|---|
| 1 | VETERINARY LEGISLATION | 42.4% |
| 2 | PUBLIC HEALTH | 37.1% |
| 3 | VETERINARY INSPECTION | 36.6% |
| 4 | FOOD CONTROL | 35.6% |
| 5 | FOOD INSPECTION | 34.8% |
| 6 | AUSTRIA | 29.5% |
| 7 | VETERINARY PRODUCT | 28.9% |
| 8 | COMMUNITY CONTROL | 28.4% |

**Table 3**. Assignment results (8 top-ranking descriptors) for the document *Food and veterinary Office mission to Austria*, found on the internet at http://europa.eu.int/comm/food/fs/inspections/vi/reports/austria/vi_rep_oste_1074-1999_en.html.

The tool does not currently attempt to recognise time expression (e.g. *5 PM*; *17:15*), date periods (e.g. *14-15 October 2004*; *in the 1960s*), incomplete dates with only one of DAY, MONTH or YEAR (e.g. *in October*; *on the third*), or named cultural festivities (e.g. *at Christmas*).

An evaluation of the tool on English texts from the *Message Understanding Conference* MUC (considering only the date expressions the tool attempts to recognise) yielded the following precision/recall values: relative dates: 86%/67%; complete dates: 100%/100%; incomplete dates: 98%/98% (for details, see Ignat 2003). The main problems regarding relative dates have since been corrected (e.g. *this may* was recognised as 'May of the reference year') so that the results are now better. The evaluation of the tool on Romanian news texts yielded similar results.

For some document types such as news articles, a list of the normalised date expressions can be a meaningful signature of the text. Together with further signatures for names, etc., documents can be described rather accurately. Following recognition, date expressions can be highlighted in text for faster retrieval (similar to place names in Figure 1). Another advantage of the application is that, once the recognised dates are normalised and stored in a database, users can search for all articles mentioning a date in a certain period, by using a simple SQL query.

### 3.2. Proper names

According to Gey (2000), 30% of content-bearing words in journalistic text are proper names such as names of people and of organisations. Friburger & Maurel (2002) showed that names recognised in text are very valuable for document similarity calculation, but say that the usage of names alone is not sufficient for this purpose. It is obvious, though, that a list of proper names can be a highly significant signature for at least journalistic text. If combined with further signatures, as proposed in this article, name lists can be very powerful.

Proper name recognition is a subject area that is very well understood and a number of named entity recognition (NER) tools are available either commercially or for research. At the JRC, we are currently using two alternative approaches to recognise people's names: (a) a PERL tool with regular expressions that identifies sequences of uppercase words as names if they are introduced or followed by cue words such as *President*, *Professor*, *teacher*, etc.; (b) the part-of-speech output of the readily trained *Tree Tagger*[11], combined with some minimalist local grammar rules. Until now, we have exploited the Tree Tagger tool only for English text, although trained Tree Tagger versions are also available for French, German and Italian.

| Spelling | Language(s) |
|---|---|
| Vladimir Putin | DA, EN, ES, IT, NO, SV |
| Vladimir Poetin | NL |
| Vladimir Poutine | FR |
| Vladimir V Putin | EN |
| Vladmir Putin | EN |
| Vladímir Putin | ES |
| Wladimir Putin | DE |
| Władimir Putin | PL |

**Table 4.** Variations of the name of the Russian President found in news texts in various languages.

The less sophisticated PERL tool misses names that are not surrounded by cue words, but it has the advantage that it is just a question of a few hours to extend it to new languages, so that we are now able to recognise names in English, French, German, Spanish, Italian, Estonian and Bulgarian.

Even within the same language and the same text, authors often use different versions of the same name. This is not only true for foreign names such as *Al Qaida* (*Al Qaeda*, *Al Kaida*, etc.), but also for known names such as *George Bush* (*George W. Bush, George Bush Jr.*, *George Walker Bush*, etc.). After having examined a number of approximative matching techniques, we decided to implement a simple letter trigram measure that allows us to recognise many monolingual and cross-lingual name variations found, as shown in Table 4. The most frequent variation is now taken as the prototypical one that is stored in the database, and all others are stored in an alias list of variations. Via an automatic lookup of the Wikipedia online encyclopaedia in various languages[12], further name variations such as Japanese ウラジーミル プーチン, Chinese 普京 and Russian Владимир Путин can be found automatically.

By using the PERL regular expressions continuously over time, a database of frequently mentioned person's names can be built up so that names can then be found in new text by using simple lookup procedures, without the need for cue words.

The result of the proper name recognition is thus a list of people's names mentioned in a given text, together with possible name variants and with information on how often the name was mentioned, both in the given text and in other texts over time. This latter frequency can be used to weight the relevance of names in a given text, using TF.IDF or a related measure, in order to down-weight frequently mentioned names such as *George Bush* and to highlight new or rarely used person names.

### 3.3. Cognates and numbers

When comparing the tokens of texts written in different languages with each other, one can frequently find some overlap. This overlap usually consists of (a) numbers in numerical form (e.g. *596*), (b) names or (c) other words that are coincidentally the same across languages (*cognates*). Cognates are normally due to common historical roots (e.g. English *finger* and *arm* vs. German *Finger* and *Arm*) or because they adapted the same loanwords (e.g. German *Computer* and Italian *computer*). These three types of identical text tokens can be exploited to contribute constructively to cross-lingual document similarity calculation. Two news articles about the same event written in English and Spanish, for instance, are likely to have a number of tokens in common, while two articles about different events are likely to have less tokens in common.

---

[11] http://www.ims.uni-stuttgart.de/projekte/corplex/TreeTagger/DecisionTreeTagger.html

[12] See various language versions at
http://en.wikipedia.org, http://de.wikipedia.org, etc.

Obviously, several limitations are linked to this approach:

(a) Number formats can differ from one language to the other, for instance due to the different usage of number separators (e.g. English 1,000.00 vs. German 1.000,00), but more often than not there is no difference (1000 is used in both languages).

(b) Names of people and places often differ from one language to the other because of different pronunciation rules (e.g. English *Al Qaeda* vs. German *Al Kaida*), or for historical reasons (e.g. English *Venice* vs. German *Venedig* vs. French *Venise*, etc.). Languages with different writing systems are much less likely to have word tokens in common, even if the pronunciation of the words is identical (e.g. Italian *Venezia* vs. Greek *Βενετία*).

(c) So-called *false friends* (words that are the same without sharing the same meaning, such as English *manifestation* and French *manifestation* or English *war* and German *war*) would cause false hits.

Many more historically related words across languages could theoretically be exploited, by writing rules that implement some historical language change phenomena. Especially the large number of European words with Greek, Latin or Germanic origin should be easy to identify: Examples include English *pharmacy* vs. French *pharmacie* and English *elephant* vs. French *elephant* vs. German *Elefant* vs. Italian *elefante*. While the benefit of the rule-based or trigram-based similarity measure has not been proven, we are already exploiting identical cognates, numbers and other identical text tokens across languages in a system for multilingual news topic tracking, as described in section 6.

## 4. Dealing with language-specific issues

From a linguistic point of view, the procedures described in the previous sections are relatively simplistic. They mainly rely on tokenisation, case information, dictionary lookup procedures, stop word lists, simple local patterns, heuristics, and statistics and Machine Learning methods operating on 'words' without part-of-speech information. Many of these procedures will work well with English texts as English has a rather poor morphology. However, this approach will be much less successful for more highly inflected languages like Hungarian or those of the Slavic language family.

It should be possible to overcome most of these phenomena with the help of good morphology tools, but these are not available to us for the large range of languages we are interested in (all twenty official EU languages and more!). As the manpower available in the JRC's Language Technology group is rather limited, as well, we had to resort, yet again, to some simple heuristics that would allow us to benefit as much as possible from the available multilingual resources and the language-independent text features while limiting the effort to a few weeks per language. With the existing applications already being set up, adding the language-specific resources for a new language takes between two and twelve weeks. Extracting the relevant terminology from the TARIC product description and preparing it for the application described in section 2.2 is rather labour-intensive so that it takes an additional estimated 12 weeks. It is clear that not all linguistic phenomena and not all languages can be dealt with, but for a large number of European languages this is sufficient to produce good and very useful text analysis applications, as described in section 6.

For the statistical EUROVOC thesaurus text classification task, experiments with Spanish have shown that, surprisingly, performance gains only approximately 2% when operating on lemmas rather than on inflected words. Furthermore, multilingual performance tests for EUROVOC descriptor assignment on eleven different languages from different language families, including German, Spanish, Finnish and Lithuanian, have shown that performance is rather uniform across the languages. Details about these experiments can be found in Pouliquen et al. (2003a).

Simple dictionary lookup procedures such as for geocoding and product recognition are, however, more sensitive to word form variations because inflected word forms such as *New Yorker* will not be found in text if the gazetteer only contains the base form *New York*. We solve this problem partially by providing language-specific regular expressions that strip potential suffixes off those uppercase words that were found in a text, but not in the place name gazetteer. For instance, if words like *Londonit*, *Frankfurdis* or *New Yorgile* are found in Estonian text, regular expressions will strip -*it* to produce *London* and will replace *dis* to *t* and *gile* to *k* in order to produce *Frankfurt* and *New York*. Together with Finnish, Estonian is known for its extremely sophisticated morphology. However, place names occur with a limited number of case endings (*in/to/from/... London*) so that 37 regular expressions cover most cases. For most languages, a much smaller number of regular expressions is needed. A small evaluation on Estonian news headlines showed that 63 out of 72 place names were recognised correctly (Recall = 87.5%). The remaining nine places were not found because either the place name was not in the database or because the suffix stripping rule was missing (about equal parts). No wrong hits occurred in the test set (precision = 100%).

It should be possible to apply the same suffix-stripping procedure to other kinds of vocabulary lists such as products, professions, etc. However, as these lists are likely to be larger and we cannot limit our search to upper case words, the lookup process should be slower and it is possible that it will produce more wrong hits.

It is not certain that for an agglutinative language like Hungarian, which can add many different types of suffixes one after the other, suffix stripping is feasible. It would be an interesting experiment to apply cascades of suffix-stripping regular expressions to see whether this helps to find place names, but the danger to get false hits due to over-stripping is big.

Further tokenisation issues arise when dealing with languages such as Chinese which do not mark word borders by a space, and compounding languages like German where (mostly) nouns can be combined to form long words. While, at least in German, expressions like *Ber-*

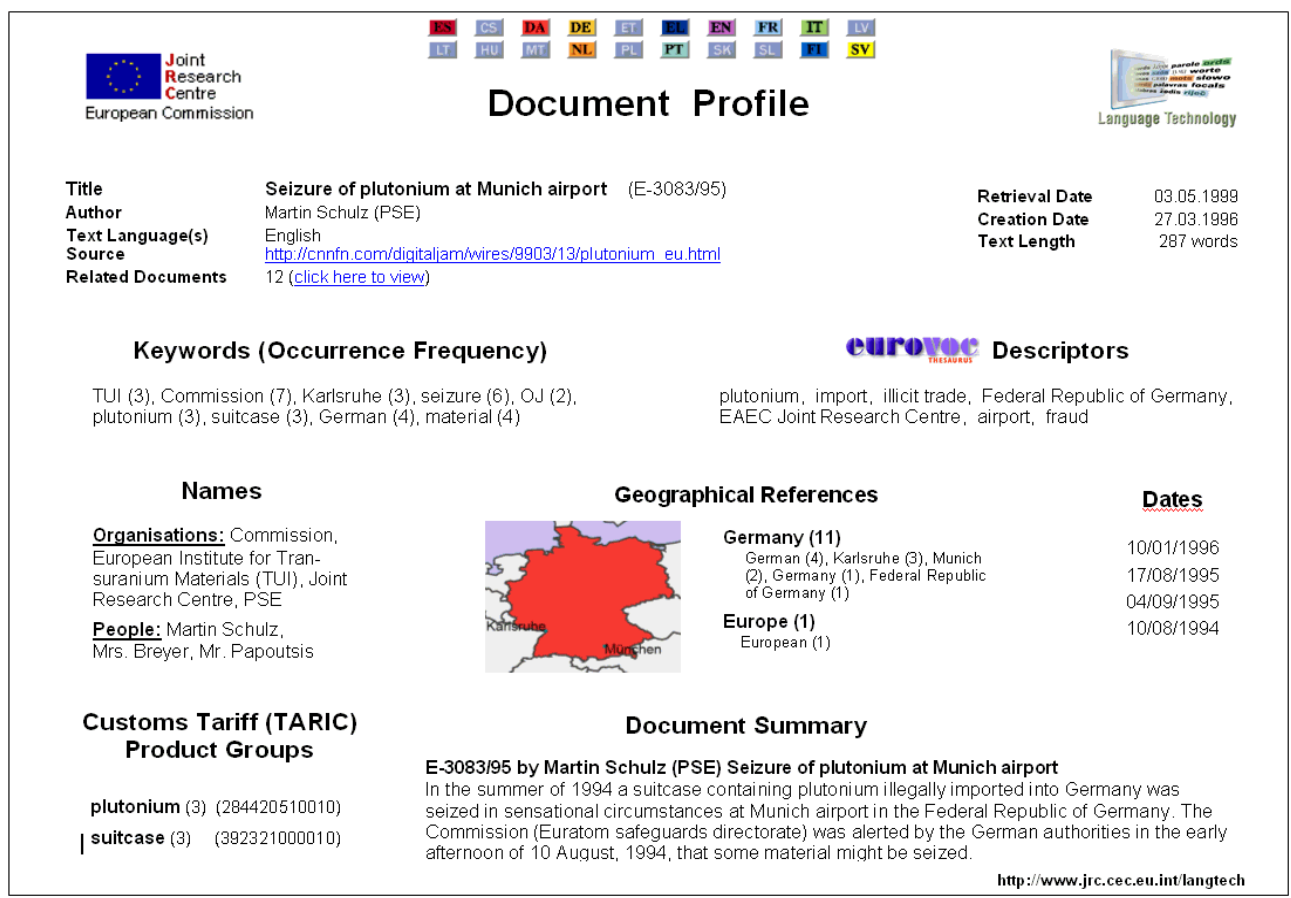

**Figure 2.** Document profile (mock-up) summarising the information extracted from documents. Entities linked to multilingual thesauri and nomenclatures can be displayed in several languages.

*liner actor* (an actor from Berlin) are not compounded (*Berliner Schauspieler*), nouns referring to products are: *Sauerstoffflaschenventilverschluss* (oxygen bottle valve closure).

For most European languages, the uppercase/lowercase distinction can be exploited when looking for the names of people or places. The same is not true for languages like Japanese, Hindi and Arabic. Furthermore, case rules even differ to some extent between languages such as English and French (e.g. *the English* vs. *les anglais*) so that rules either have to be adapted specifically to each language or lower recall has to be accepted when looking only at uppercase words.

## 5. Language-independent procedures and applications

In the highly multilingual setting of the set of applications discussed in this article, language-independent text analysis procedures are very useful. We currently use the following applications:

(a) An automatic language guessing tool using letter bigram and trigram statistics, that has currently been trained for 25 languages.
(b) A keyword extraction tool that identifies the statistically most salient words and their relative importance (their *keyness*) by comparing the word frequency in the text with an average word frequency as found in large reference corpora. While we use the log-likelihood formula to extract and rank the words, other formulae like TF.IDF are possible alternatives. A list of stop words can be used to stop some words from being identified as keywords that are low in semantic content or that are meaningless when being out of context. A ranked list of keywords for a document is a good vector space representation of this document.
(c) A tool to measure the similarity between two documents by calculating the cosine or another similarity measure between the vector space representations of two documents. Monolingually, the list of extracted keywords and their keyness can be used as input. For cross-lingual similarity calculation, features like the ones discussed in this article can be fed to the system.
(d) This document similarity measure can be used for a number of applications, including hierarchical unsupervised document clustering, classification and query-by-example document retrieval.

Further applications that can be based on language-independent methods are automatic document summarisation by extracting the most relevant sentences (e.g. those containing most keywords), and the generation of document maps. Document maps such as Kohonen maps are two-dimensional representations of the multi-dimensional document space that can be useful to get a first overview of the main contents of a large document collection or to navigate in the document collection.

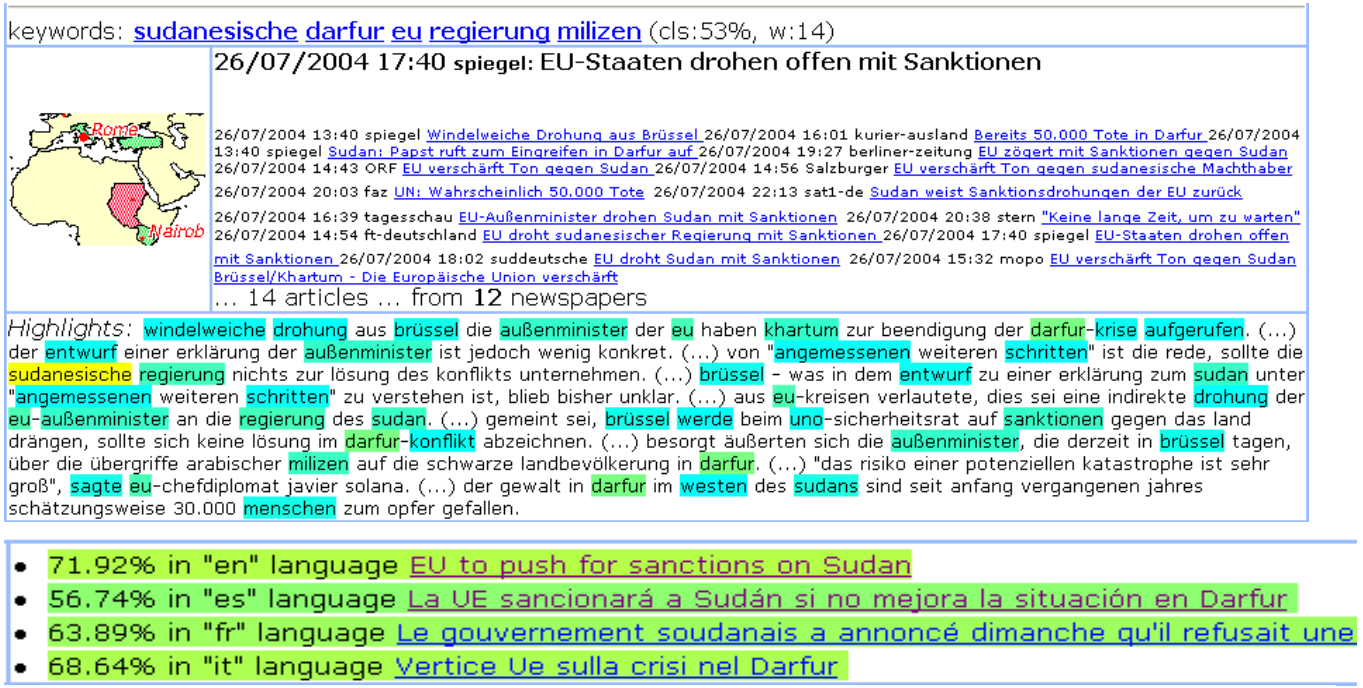

**Figure 3.** German news automatically identified as being about the same subject, together with the title of the most representative news article, the keywords for this cluster and a map showing the place names mentioned in the cluster. The links below lead to the corresponding news article cluster in English, Spanish, French and Italian.

## 6. Applications

At the JRC, we combine applications based on the language-independent algorithms listed in section 5 with the information extracted according to the procedures described in sections 2 and 3. In spite of the relatively shallow linguistic processing, we were able to produce applications that are being used as regular in-house services and for the ad-hoc analysis of document collections given to us by various users.

Once entities such as dates, names or products have been identified, they can be highlighted in text in different colours to allow users to find them quickly, as shown in Figure 1. For foreign language text, the entity can be displayed in another language to give users information about a text that they might not otherwise understand (*cross-lingual information access*). The various information aspects (products, places, keywords, etc.) extracted from unrestricted and unstructured text can also be displayed together to provide users with sort of a document profile, as shown in Figure 2. Those information aspects that are linked to multilingual nomenclatures, gazetteers and thesauri can furthermore be displayed in languages other than the document language.

The structured meta-information is stored in a database to enable users to search document collections by using this meta-data as features. This makes it possible, for instance, to search for all documents mentioning tobacco products, making reference to Turkey and mentioning a date in the range 1.01.2003 and 31.03.2003.

When the reference of geographical place names has been identified unambiguously, i.e. when we have identified latitude and longitude of the places, it is easy to create a map showing the geographical coverage of a document, of a cluster of documents or of a whole document collection. Figure 3 shows a small map with those geographical places highlighted that were mentioned in a cluster of news articles about the same subject. It also shows how the clustering of news represented by their automatically identified keywords successfully identifies all those articles that talk about the same event (in Figure 3, it is the discovery of our solar system's tenth planet, Sedna, in March 2004). The vector space representation of the whole cluster can be compared to that of each individual article, by calculating the cosine, so that the article whose representation is closest to the centroid of the cluster's representation can be chosen as the most typical article whose title can be chosen as the cluster title.

Figure 3 also shows how cross-lingual links between a cluster and the news clusters in other languages can be established successfully by using the multilingual nomenclatures, thesauri and gazetteers as an interlingua. The JRC's cross-lingual news tracking system (Pouliquen 2004b) represents each cluster by three different vectors. When comparing this document representation with those of clusters in other languages, each of the three vectors contributes with a different weight to the overall similarity between the clusters of documents written in different languages, as described in Pouliquen (2004b). Another usage of the cross-lingual document similarity calculation is the automatic compilation of collections of parallel (or comparable) texts to train and test information extraction or Machine Translation software. When testing the document similarity calculation based only on the EUROVOC descriptor vector representation of 820 English documents and their Spanish translations (Pouliquen et al. 2003b), we found that in 90.61% of cases, the Spanish translation was successfully found as being the most similar Spanish document for a given English document. When adding information about the length of texts to exploit the fact that translations should have a similar length to the original document, the result increased to 96.83%. This result shows that processes to map documents onto a multilingual thesaurus can lead to extremely powerful applications. Cross-lingual document similarity calculation is also an essential ingredient for cross-lingual document plagiarism detection, an application for which, to our knowledge, no solutions have been proposed to date.

## 7. Conclusion

The intention of this article was to describe how multilingual knowledge sources such as gazetteers, vocabulary lists, nomenclatures and thesauri, as well as language-independent text features such as dates, can be exploited for information extraction tasks, to provide cross-lingual information access and to calculate cross-lingual document similarity, which itself is a basic ingredient for many more text analysis applications. We furthermore wanted to show how relatively naïve text analysis tools can be helpful to develop powerful text analysis applications for many different languages with rather little effort, once the methodology has been decided on and the tools have been set up. At the JRC, we have already developed the language-specific resources for a number of European languages and we are currently making an effort to extend this tool set to all twenty official languages of the European Union. While we have no doubt that it is possible to produce better results with more thorough linguistic methods, such labour-intensive language-specific work is not an option for our small team whose aim it is to work on 20 or more languages. Instead, we exploit existing multilingual lexical resources (even if they had not initially been developed for machine use) and language-independent text features, and we make use of Machine Learning techniques, statistical methods and heuristics. We believe to have shown that this approach can lead to good results and that it is even possible to produce working versions of novel applications such as cross-lingual news topic tracking using an interlingua document representation. The effort required to develop the language-specific resources for a new language ranges between one week and three months for the applications we are currently using. Extracting and developing TARIC product nomenclature terms is a comparatively labour-intensive task that requires an additional estimated two to three months. In order to extend the current tool set to new languages and applications, the JRC is actively seeking collaborators such as mother tongue students who would join us as trainees.

Individual applications out of the set presented in this paper have been tested and proven, including date and place name recognition, EUROVOC thesaurus descriptor assignment, monolingual news clustering and news topic tracking, and cross-lingual news topic tracking. A number of other applications presented here still need to be evaluated formally. Furthermore, it would be useful to carry out a thorough one-by-one evaluation of the effectiveness of each of the text features presented here, and of their relative impact for cross-lingual document similarity calculation.

The JRC can share tools and resources with non-commercial entities if they are not bound by copyrights owned by other organisations. The JRC is furthermore interested in collaborations yielding more language resources, especially for the new EU languages.

## Acknowledgements
Many people have contributed to developing the tool set described in this paper and to developing and evaluating the language-specific resources for various languages. We would particularly like to thank Laima Norvilienè (born Cekyte) and Irina Temnikova for their help with the product recognition tool, Victoria Fernandez Mera, Elisabet Lindkvist Michailaki and Arturo Montejo-Ráez for their help regarding the EUROVOC thesaurus indexing application, Marco Kimler for his refinement of the geo-coding tool, and Emilia Käsper, Ippolita Valerio, Tom de Groeve, Victoria Fernandez Mera, Tomaž Erjavec, Christian Gold and Irina Temnikova for their help in creating language-specific resources for Estonian, Italian, Dutch, Spanish, Slovene, German, Bulgarian and Russian. We would also like to thank the JRC's Web Technology team for providing us with the multilingual news collection to develop and test many of the applications described here.